\pdfoutput=1

\documentclass{article}

\usepackage{times}
\usepackage{changes}
\usepackage{graphicx} 
\usepackage{natbib}
\usepackage{subcaption}
\usepackage{algorithm}
\usepackage{algorithmic}
\usepackage{hyperref}
\usepackage{amsmath, amssymb, amsthm}
\usepackage{xcolor}
\usepackage{enumitem}
\usepackage[accepted]{icml2014}

\renewcommand\emph[1]{\textit{#1}}

\theoremstyle{definition}



\icmltitlerunning{A Sober Look at Spectral Learning}

\begin{document} 

\twocolumn[
\icmltitle{A Sober Look at Spectral Learning}

\icmlauthor{Han Zhao}{han.zhao@uwaterloo.ca}
\icmlauthor{Pascal Poupart}{ppoupart@uwaterloo.ca}
\icmladdress{Cheriton School of Computer Science, University of Waterloo, Waterloo, Ontario, Canada}

\icmlkeywords{Spectral Learning, Hidden Markov Model, Maximum Likelihood, Model Selection}

\vskip 0.3in
]

\begin{abstract}
Spectral learning recently generated lots of excitement in machine learning, largely because it is the first known method to produce consistent estimates (under suitable conditions) for several latent variable models.  In contrast, maximum likelihood estimates may get trapped in local optima due to the non-convex nature of the likelihood function of latent variable models.  In this paper, we do an empirical evaluation of spectral learning (SL) and expectation maximization (EM), which reveals an important gap between the theory and the practice. First, SL often leads to negative probabilities. Second, EM often yields better estimates than spectral learning and it does not seem to get stuck in local optima. We discuss how the rank of the model parameters and the amount of training data can yield negative probabilities.  We also question the common belief that maximum likelihood estimators are necessarily inconsistent.  
\end{abstract}

\section{Introduction}
Spectral Learning is a general approach that uses spectral decompositions (e.g., singular value decomposition and tensor decomposition) for parameter estimation based on the method of moments~\cite{Hsu2012,Parikh2011,Anandkumar,Anandkumar2012a,Anandkumar2012}. Spectral learning has generated a lot of excitement in recent years due to its performance guarantees in latent variable models.  The presence of discrete latent variables generally leads to a non-concave log-likelihood function, which is problematic for maximum likelihood estimators. Spectral learning is the first known method to be consistent (under suitable conditions) for several latent variable models including mixtures of Gaussians (MoGs), hidden Markov models (HMMs) and latent Dirichlet allocation (LDA). Furthermore, finite sample bounds guarantee that the approach will find nearly optimal parameters or make nearly optimal predictions with high probability given a sufficient amount of training data~\cite{Hsu2012}. 

We report some experiments that suggest an important gap between the theory and the practice. Despite its theoretical guarantees, spectral learning often generates negative probabilities. This is an important issue that has received little attention so far. We show some empirical results that suggest that a poor choice of the rank of the model parameters and insufficient training data increase the likelihood of negative probabilities. 
We also investigate how well spectral learning performs in comparison to common approaches such as EM that do not enjoy the same theoretical guarantees. Interestingly, even though EM is subject to local optima and spectral learning is not, EM often outperforms spectral learning. Contrary to the common belief, we suggest that EM may be consistent in several settings. We discuss two situations under the assumption that the observation space is finite. When the true parameters are identifiable, increasing the amount of data often leads to a unimodal (though still non-concave) likelihood function, which explains why maximum likelihood estimators do not suffer from local optima. When the true parameters are unidentifiable (i.e., several equivalent solutions), the likelihood function remains multimodal, but if all the peaks of the likelihood function are at equivalent solutions, maximum likelihood estimators do not suffer from local optima. We also discuss two advantages of maximum likelihood estimators over spectral learning: a) maximum likelihood is a better objective to optimize than moment consistency and b) the data efficiency of maximum likelihood tends to be higher since it uses all empirical moments of the data (not just a few low order moments).

\section{Spectral Learning for HMMs}
Consider an HMM described as follows. Let $x_1,x_2,x_3,\ldots$ denote a sequence of discrete observations where $x_t\in [n] = \{1,\ldots,n\}$ is the observation at time step $t$, and $h_1,h_2,h_3,\ldots$ denotes a sequence of hidden states where $h_t\in [m] = \{1,\ldots,m\}$ is the hidden state at time step $t$. The parameters of an HMM are $(\pi, T, O)$ where $\pi\in\mathbb{R}^m$ is the initial state distribution, $T\in\mathbb{R}^{m\times m}$ is the transition matrix and $O\in\mathbb{R}^{n\times m}$ is the observation matrix. More specifically, we have $\Pr(h_1 = i) = \pi_i$, 
$\Pr(h_{t+1} = i | h_t = j) = T_{ij}$ and $\Pr(x_t = i|h_t=j) = O_{ij}$.
Based on $(\pi, T, O)$, we define an observable operator $A_x = T\text{diag}(O_{x,1},\ldots, O_{x,m})\in\mathbb{R}^{m\times m}$ for each observation $x\in[n]$. The joint probability of an observation sequence of length $t$ can be computed based on these operators as follows:
\begin{equation}
\Pr(x_1,\ldots,x_t) = \mathbf{1}_m^T A_{x_t}\ldots A_{x_1}\pi
\end{equation}

\citet{Hsu2012} proposed a spectral algorithm called LearnHMM to estimate a transformed set of operators based on some low order empirical moments of the data.  The following moment matrices are estimated from the data:
 $$\begin{array}{l l}
P_1\in\mathbb{R}^n, &  [P_1]_i = \Pr(x_1 = i)\\
P_{2,1}\in\mathbb{R}^{n\times n}, & [P_{2,1}]_{ij} = \Pr(x_2 = i, x_1 = j)\\
P_{3,x,1}\in\mathbb{R}^{n\times n}, &  [P_{3,x,1}]_{ij} = \Pr(x_3 = i, x_2 = x, x_1 = j)
\end{array}$$
LearnHMM requires a matrix $U\in\mathbb{R}^{n\times m}$ such that $U^TO$ is invertible. It is often chosen to be the first $m$ left singular vectors that preserve the range of $O$. The following operators are then computed:
\begin{eqnarray}\label{equ:operators}
b_1 & = & U^TP_1\nonumber\\
b^T_\infty & = & P_1^T(U^TP_{2,1})^+\nonumber\\
B_x & = & U^TP_{3,x,1}(U^T P_{2,1})^+  \;\; \forall x \in[n]
\end{eqnarray}
If $T$ and $O$ are of rank $m$ and $\pi$ is element-wise positive, it can be shown that 
$$\Pr(x_1,\ldots,x_t) = b_\infty^T B_{x_t}\ldots B_{x_1}b_1$$
The classic parameters $(\pi,O,T)$ can also be recovered from the operators~\cite{Hsu2012}.

In practice, since we do not know the exact moments, we obtain approximate moment matrices $\widehat{P}_1$, $\widehat{P}_{2,1}$, $\widehat{P}_{3,x,1}$ from the data and approximate operators $\widehat{b}_1$, $\widehat{b}^T_\infty$, $\widehat{B}_x$.
\citet{Hsu2012} proved that joint probability estimates are consistent in the sense that 
\begin{equation}
\lim_{N\rightarrow\infty}\sum_{x_1,\ldots,x_t}|\Pr(x_1,\ldots,x_t)-\widehat{\Pr}(x_1,\ldots,x_t)| = 0
\end{equation} 
where $N$ is the sample size. They also showed that $\forall\epsilon > 0$, the sample size needed to get an $\epsilon$-bound on the estimate is polynomial in $t$ and $m$.  Several extensions and variants of this approach have been proposed for many latent variable models~\cite{Parikh2012,Parikh2011,Anandkumar,Anandkumar2012a}.

\section{Negative Probabilities}
\begin{figure*}[htbp]
\centering
	\begin{subfigure}[b]{0.28\textwidth}
	\centering
		\includegraphics[scale=0.35]{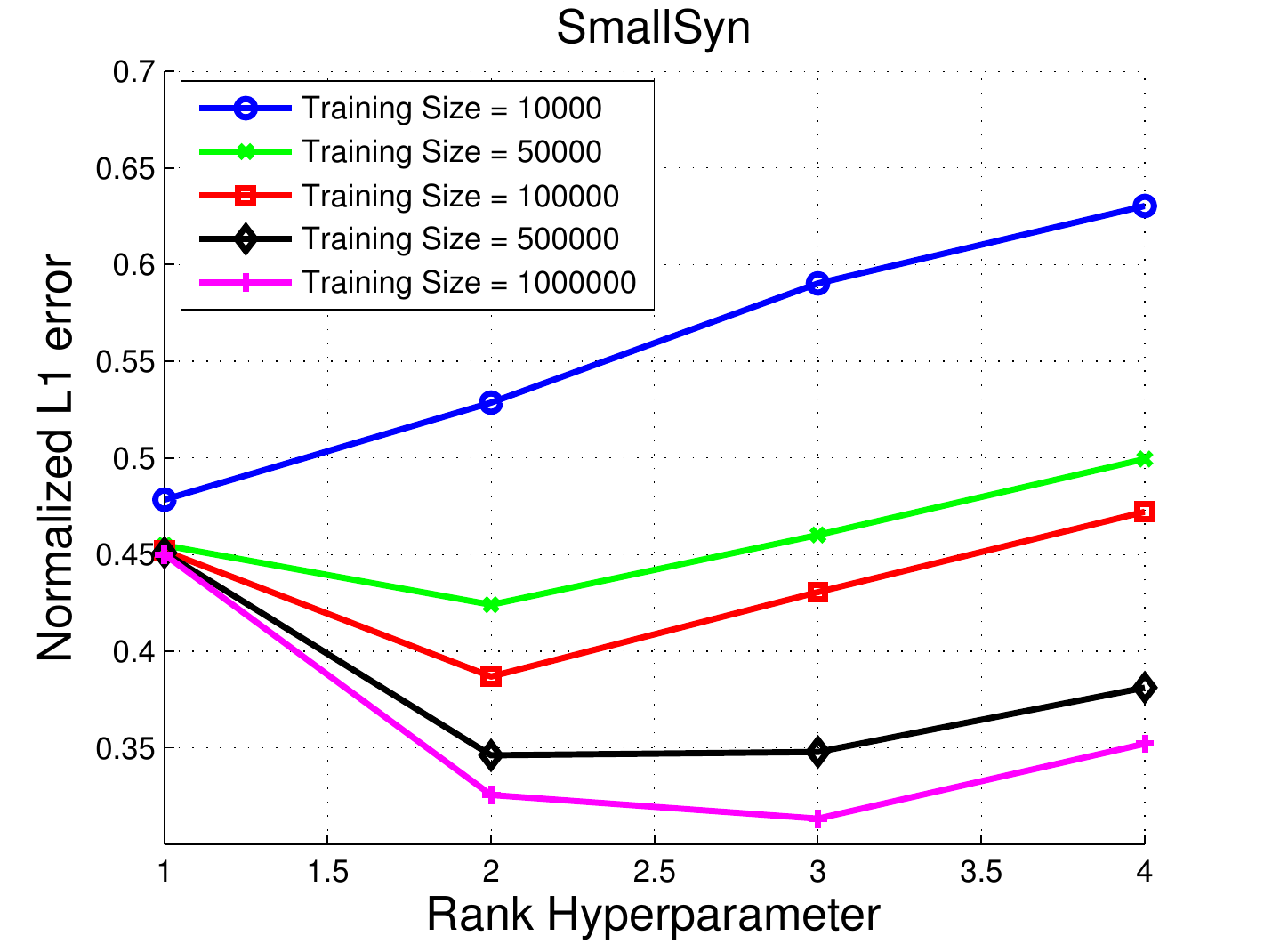}
		\caption{}\label{fig:small-syn-l1}
	\end{subfigure}
	\hfill
	\begin{subfigure}[b]{0.28\textwidth}
	\centering
		\includegraphics[scale=0.35]{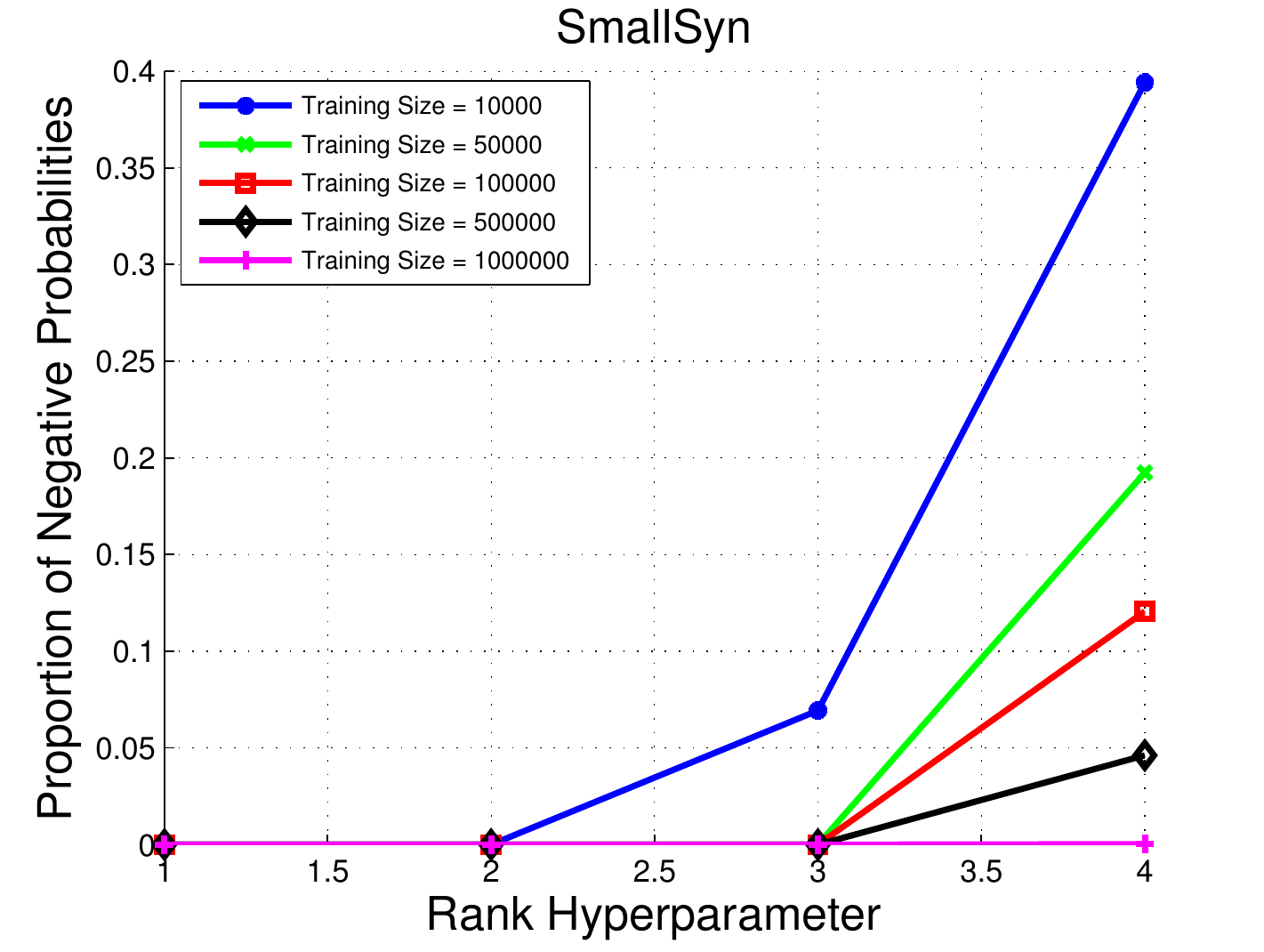}
		\caption{}\label{fig:small-syn-neg}
	\end{subfigure}
	\hfill
	\begin{subfigure}[b]{0.28\textwidth}
	\centering
		\includegraphics[scale=0.35]{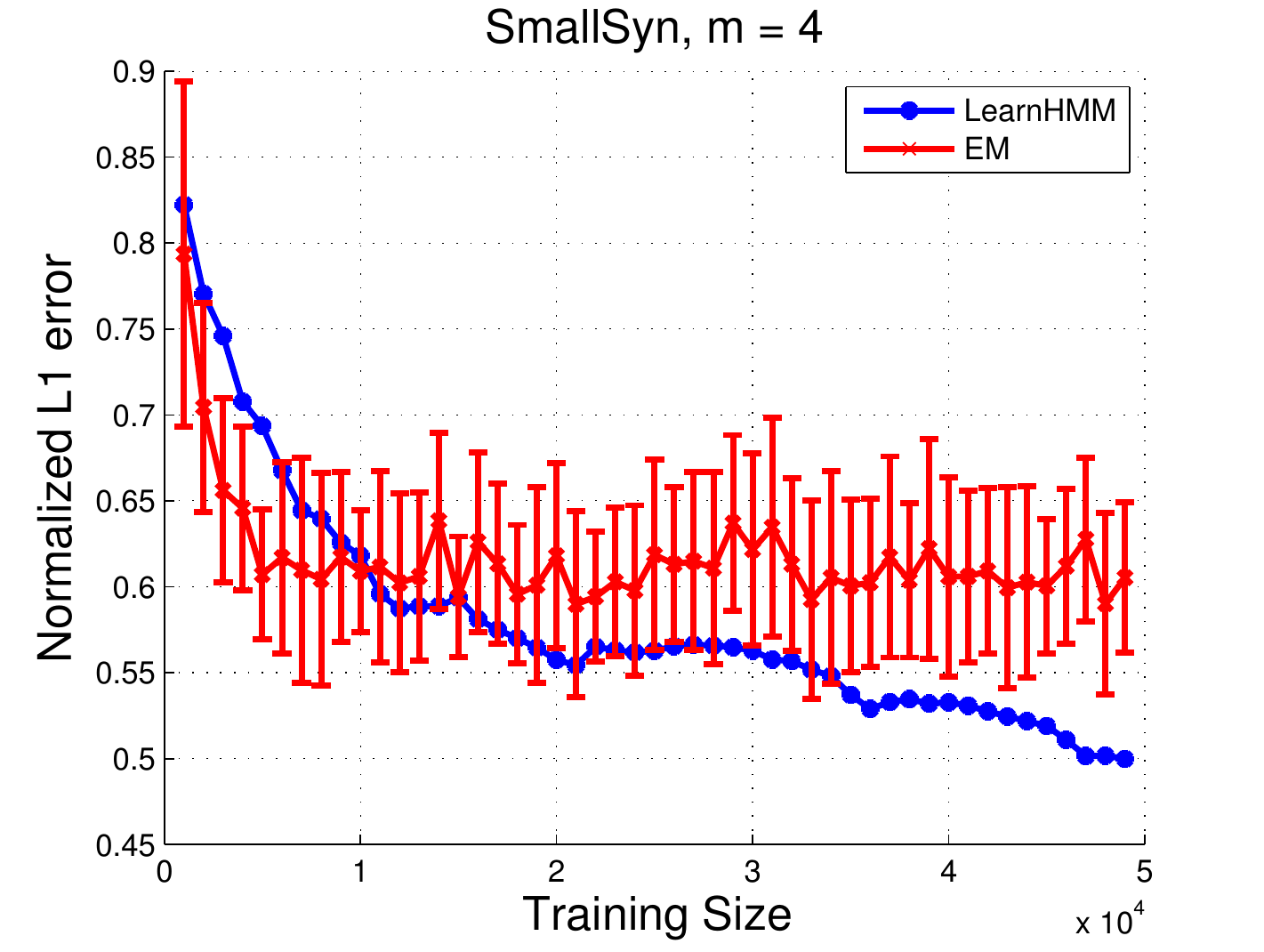}
		\caption{}\label{fig:sl-em-small}
	\end{subfigure}
	\hfill
	\begin{subfigure}[b]{0.28\textwidth}
	\centering
		\includegraphics[scale=0.35]{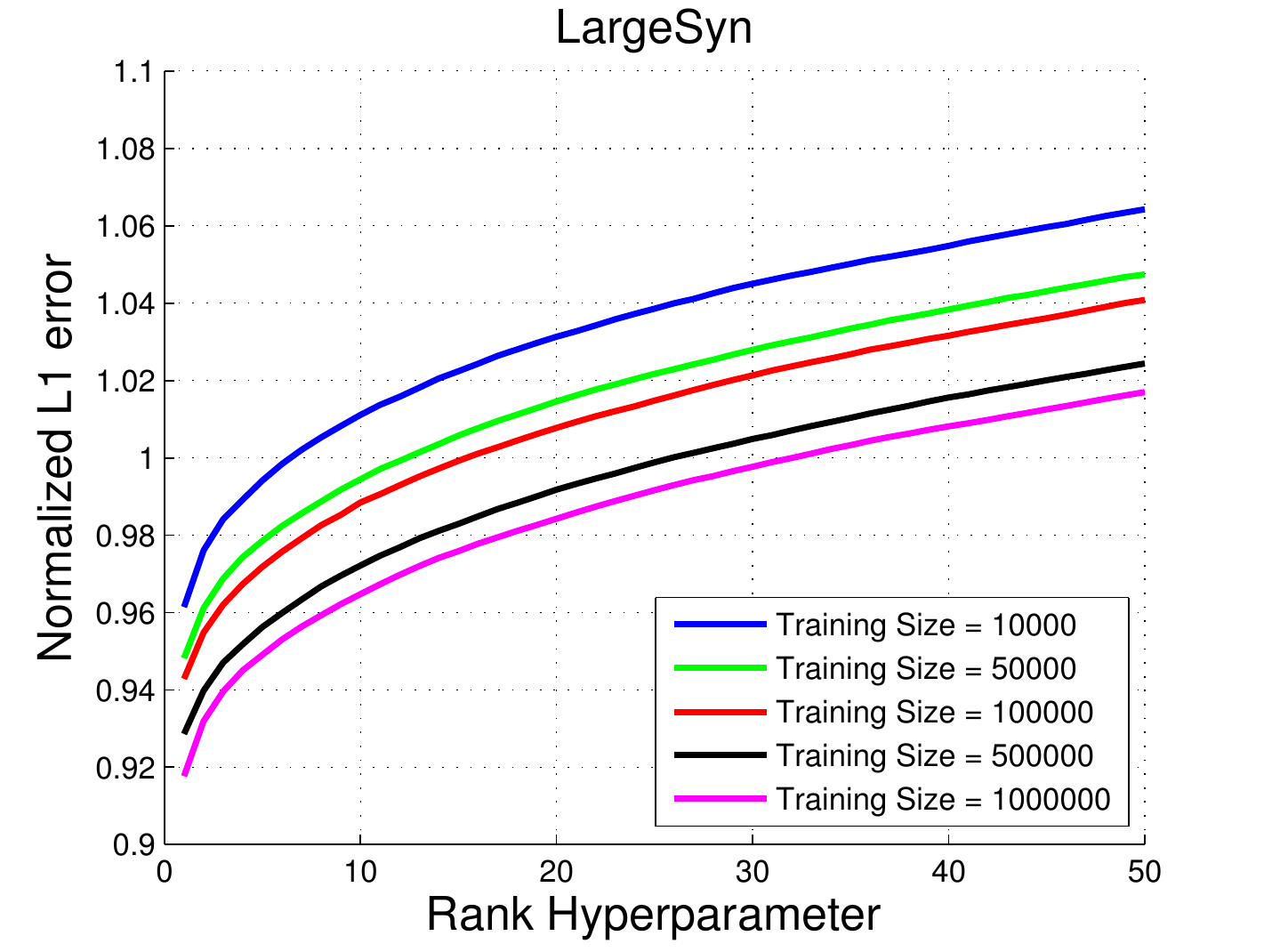}
		\caption{}\label{fig:large-syn-l1}
	\end{subfigure}
	\hfill
	\begin{subfigure}[b]{0.28\textwidth}
	\centering
		\includegraphics[scale=0.35]{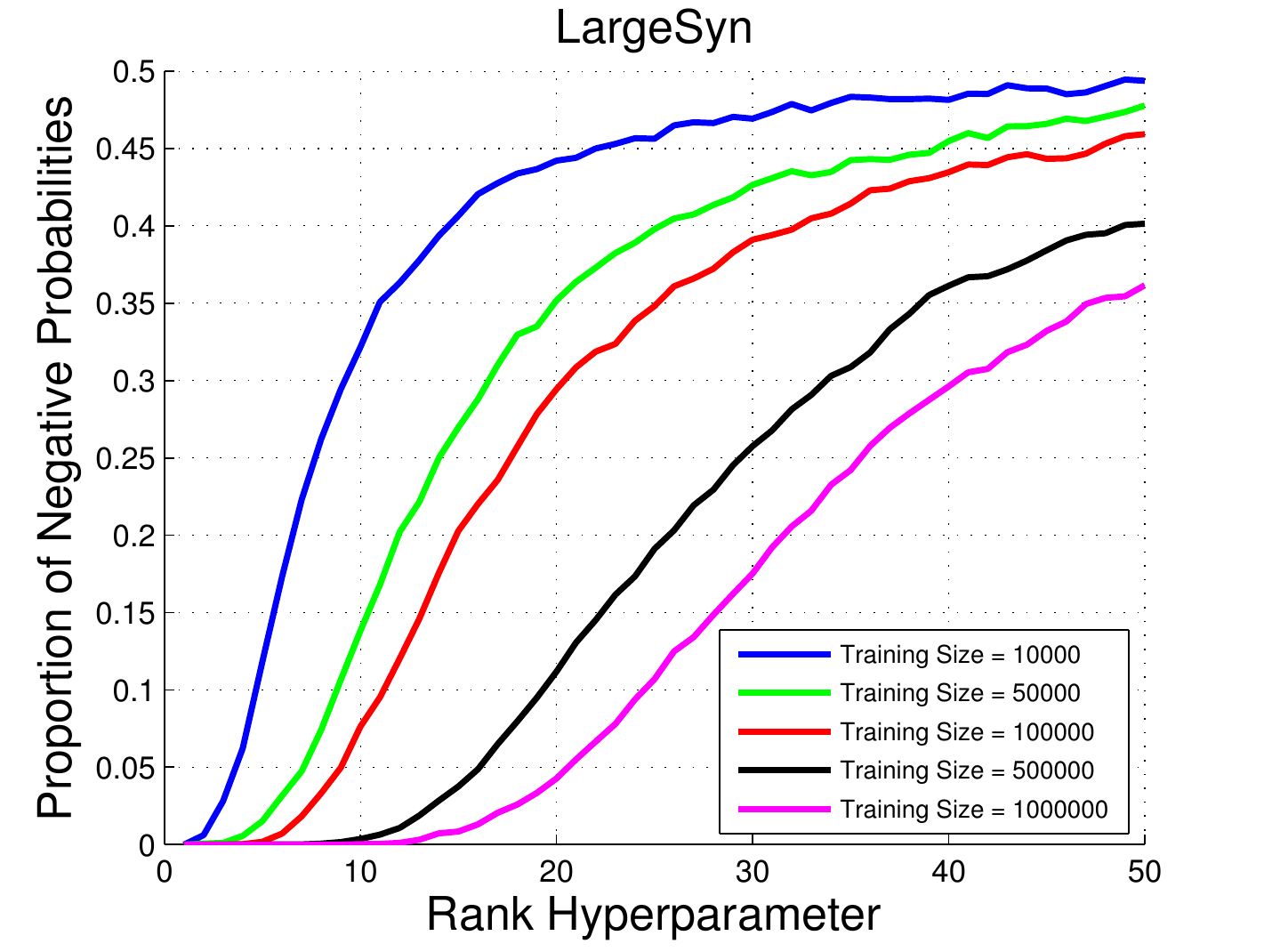}
		\caption{}\label{fig:large-syn-neg}
	\end{subfigure}
	\hfill
	\begin{subfigure}[b]{0.28\textwidth}
	\centering
		\includegraphics[scale=0.35]{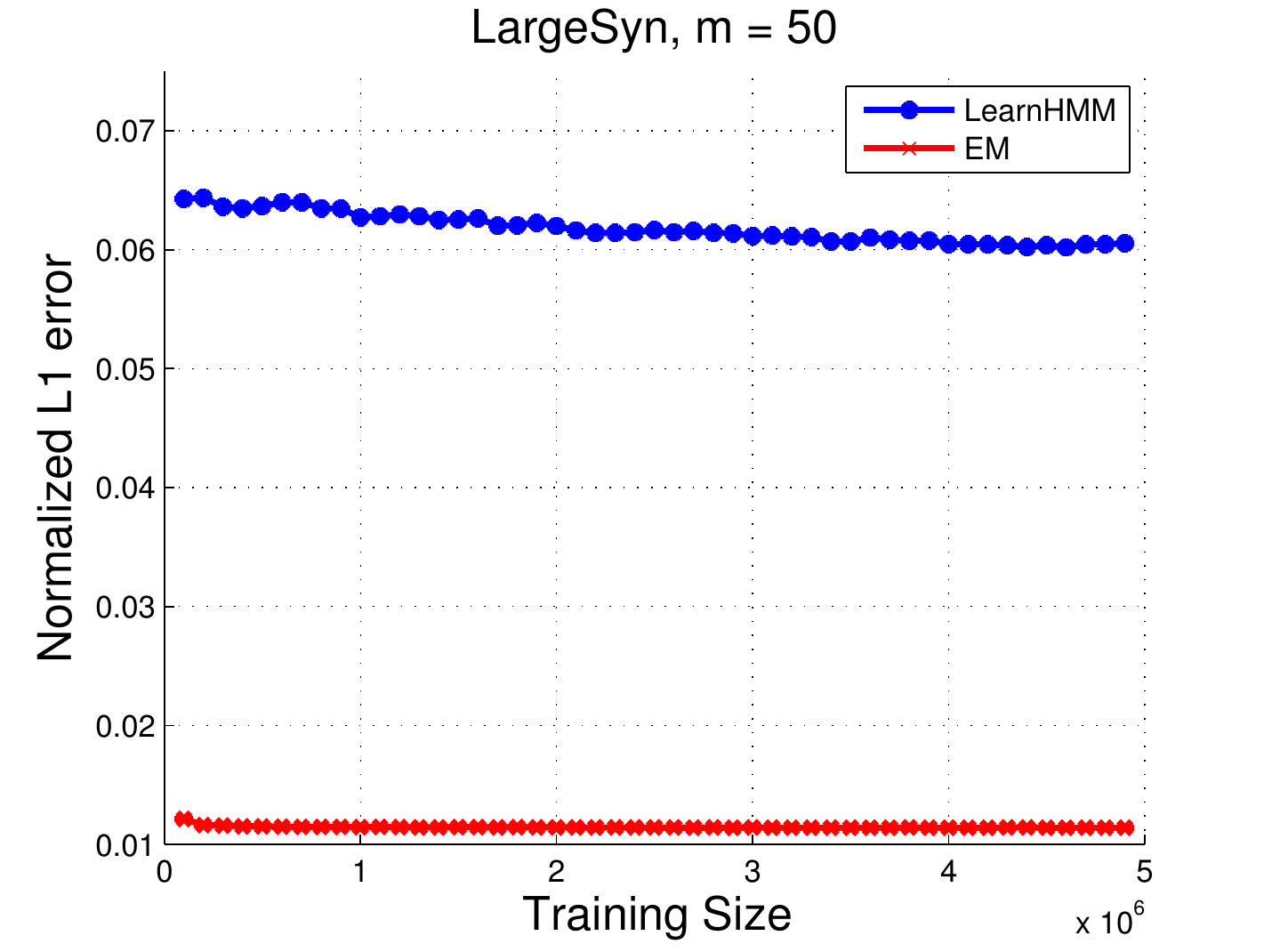}
		\caption{}\label{fig:sl-em-large}
	\end{subfigure}
\caption{Results for LearnHMM on two data sets (Fig.\ref{fig:small-syn-l1},\ref{fig:small-syn-neg},\ref{fig:large-syn-l1},\ref{fig:large-syn-neg}) and comparing LearnHMM with EM (Fig.\ref{fig:sl-em-small},\ref{fig:sl-em-large}).}
\label{fig:results}
\end{figure*}

We implemented LearnHMM and tested it on small and large synthetic discrete HMMs.  The small HMM has 4 hidden states, 8 observations and the test set consists of 4096 observation sequences of length 4. The large HMM has 50 hidden states, 100 observations and a test set of 10,000 observation sequences of length 50. Fig.~\ref{fig:small-syn-l1} and Fig.~\ref{fig:large-syn-l1} show the normalized $L_1$ error when estimating the probability of the test sequences as we vary the amount of training data and the rank hyperparameter $m$. The normalized $L_1$ error is defined as follows:
$$L_1 = \sum_{(x_1,\ldots,x_t)\in \mathcal{T}}|\Pr(x_1,\ldots,x_t)-\widehat{\Pr}(x_1,\ldots,x_t)|^{\frac{1}{t}}$$
where  $\mathcal{T}$ is the set of test sequences. We also report the proportion of negative probabilities 
$$\text{NEG\_PROP} = \frac{|\{\widehat{\Pr}(x_1,\ldots,x_t)<0~|~(x_1,\ldots,x_t)\in\mathcal{T}\}|}{|\mathcal{T}|}$$
computed by LearnHMM in Fig.~\ref{fig:small-syn-neg} and Fig.~\ref{fig:large-syn-neg}. Negative probabilities are an important problem as they occur frequently. 
Increasing the amount of data and choosing a more accurate rank parameter tends to decrease the $L_1$ error and the proportion of negative probabilities.

Negative probabilities do not invalidate the theoretical guarantees of spectral learning. They simply reflect the fact that the theoretical guarantees are expressed in terms of bounds on \textit{additive} error (see Theorem 6 in \citet{Hsu2012}). When the true probability is close to 0 and the bound is loose, it may guarantee that the estimated probability is in some interval that is partly negative. The problem of negative probabilities is well-known in the literature on observable operator models and was acknowledged by~\citet{boots2011closing} who rounded up all negative outputs to a number slightly above zero followed by normalization.  
In some spectral learning algorithms such as Excess Correlation Analysis for latent Dirichlet allocation~\cite{Anandkumar}, the parameters are estimated up to a sign. This means that an exact estimate of a distribution will normally be all positive or all negative and the sign can be flipped in the case of an entirely negative distribution.  However, since the parameters are estimated approximately, the sign of the probabilities will often be mixed. 
It is not clear anymore whether the sign should be flipped. A simple heuristic consists of adding the probabilities of all outcomes and if the sum is negative, then flip the sign of all probabilities.  After that, the negative probabilities can be rounded up to a number slightly higher than 0 followed by normalization.  Here, there is a risk that the sign of the probabilities will be flipped when it should not.  If most of the mass is negative due to the approximate nature of spectral learning, then the sign should not be flipped. 
Those heuristics will ensure that the final probabilities are positive, but they may increase the additive error. Spectral learning (with those heuristics) remains consistent in the limit, but the finite sample bounds need to be revised (this is an open problem).

Can we modify spectral learning to ensure that all probabilities are non-negative? The root of the problem is that spectral learning implicitly solves a system of non-linear equations without restricting the space to non-negative solutions. 
Could we simply add additional constraints to ensure non-negativity?   We conjecture that it will be NP-hard.  Consider the problem of matrix factorization (i.e., find matrices A and B such that C=AB). If the entries of A and B may be any real number, then a solution can be found in polynomial time by singular value decomposition. However, if we want A and B to be non-negative then this becomes a problem of non-negative matrix factorization, which is NP-hard~\cite{vavasis2009complexity}. Similarly, spectral learning finds operators (from which transition and observation matrices can be recovered) by singular value decomposition in polynomial time. If we add non-negativity constraints for the resulting transition and observation matrices, we conjecture that the problem will become NP-hard.

\section{Empirical Comparison with EM}
We compared empirically spectral learning to expectation maximization (EM) on synthetic HMMs. The theory suggests that spectral learning should perform better since it is consistent while EM is subject to local optima, but the results are mixed.  On the small synthetic HMM (Fig.~\ref{fig:sl-em-small}), with the true rank and sufficient data, spectral learning outperforms EM, but on the large synthetic HMM (Fig.~\ref{fig:sl-em-large}), EM outperforms spectral learning. The amount of training data was the same for both problems.  We suspect that the amount of training data was insufficient for spectral learning to estimate reasonable operators for the larger model. Furthermore, since spectral learning inverts a matrix to recover a similarity transform of the observable operators, it is unstable and sensitive to noise. We also noticed a large amount of negative probabilities.  In general, spectral learning is very sensitive to the amount of data and the rank parameter as discussed in the previous section.  

We also note that spectral learning does not optimize any desirable objective.  Since it implicitly solves a non-linear system of equations induced by moment matching, if the moments matrices are too approximate, the resulting operators may be far from those that produced the data. If the system of equations is highly sensitive to perturbations, then spectral learning may yield terrible results. In contrast, EM directly maximizes the likelihood of the data.  So even when there is little data it will find parameters that are likely to generate the data.  The main issue with small datasets for EM is overfitting. We did not use regularization to mitigate overfitting in this experiment.  

Another difference between spectral learning and EM is the information from the data that is used in training.  Spectral learning does not use the raw training data.  It uses only the first few empirical moments, which can be viewed as insufficient statistics.  In contrast, EM trains with the raw data and therefore implicitly takes into account all the empirical moments including the higher order moments that spectral learning ignores. 

\section{Local Optima}
\begin{figure}[htbp]
\centering
\includegraphics[scale=0.35]{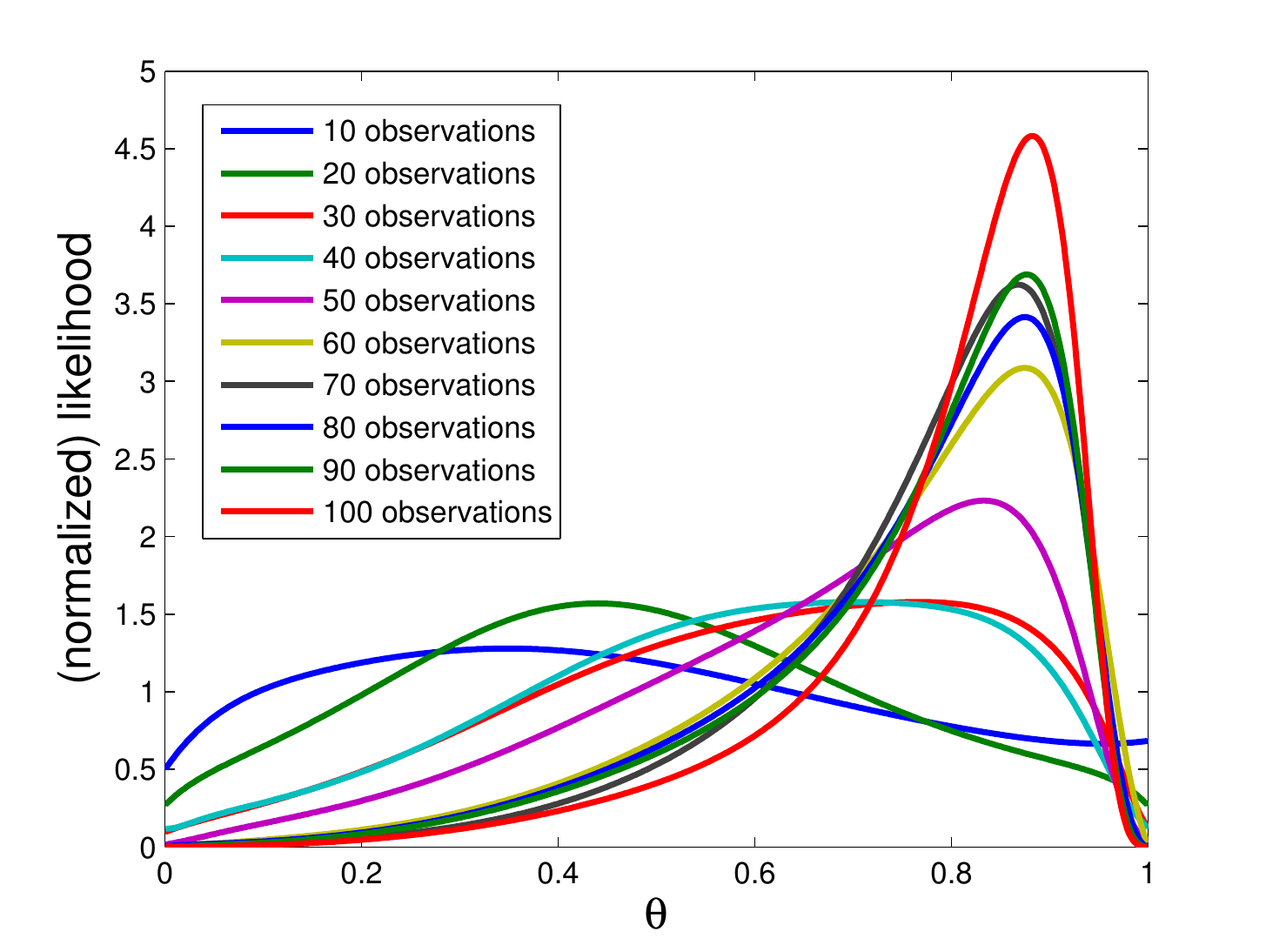}
\caption{Unnormalized likelihood curves $\Pr(\theta|data)$.}
\label{fig:unimodal}
\end{figure}
We were surprised by the fact that EM performed as well as it did since it may get stuck in arbitrarily bad local optima. While EM produced different results for different random restarts as shown by the error bars in Fig.~\ref{fig:sl-em-small} and~\ref{fig:sl-em-large}, the results were generally quite good and consistent. The variation for different random results can be explained by (minor) local optima or different amounts of overfitting.  To investigate this further we constructed a single-parameter HMM for which we can visualize the likelihood function.  This HMM has 2 states and 2 observations.  We assume that the observation distribution is known and fixed ($\Pr(x_i|h_i) = 0.7 \; \forall i$) while the transition distribution is symmetric with a single parameter $\theta = \Pr(h_{i+1}=n|h_i=n) \; \forall n,i$ indicating the probability that the current state remains unchanged. The likelihood function can be computed analytically for discrete latent variable models since it consists of an unnormalized mixture of Dirichlets. However, this mixture has one Dirichlet component per joint assignment of the latent variables. For a sequence of $t$ time steps, this would yield an exponentially large mixture in $t$.  However, when there is only one parameter $\theta$, several mixture components can be collapsed together and the number of \emph{different} components in the mixture grows quadratically in $t$. Figure~\ref{fig:unimodal} shows the analytical likelihood function for data sequences of increasing length. Each curve is a mixture of Dirichlets corresponding to the likelihood function for observation sequences of different length. Since the likelihood functions are (unnormalized) mixtures of Dirichlets, we expect to see multimodal curves, but most of the curves in Fig.~\ref{fig:unimodal} are unimodal. This means that maximum likelihood estimators such as EM will perform very well.  The fact that mixtures of Dirichlets tend to form unimodal curves as we increase the amount of data can be explained by the consistency of Bayesian learning~\cite{casella1990statistical}. When we start with a uniform prior, the posterior in Bayesian learning is the likelihood function. Since Bayesian learning is consistent for discrete observation models, the posterior converges to a Dirac distribution in the limit as long as the true parameters are identifiable (i.e., unique solution)~\cite{casella1990statistical}. Hence the likelihood function converges to a Dirac distribution too and we conjecture that EM is consistent in this setting.  

\begin{figure}[bht]
\centering
\includegraphics[scale=0.35]{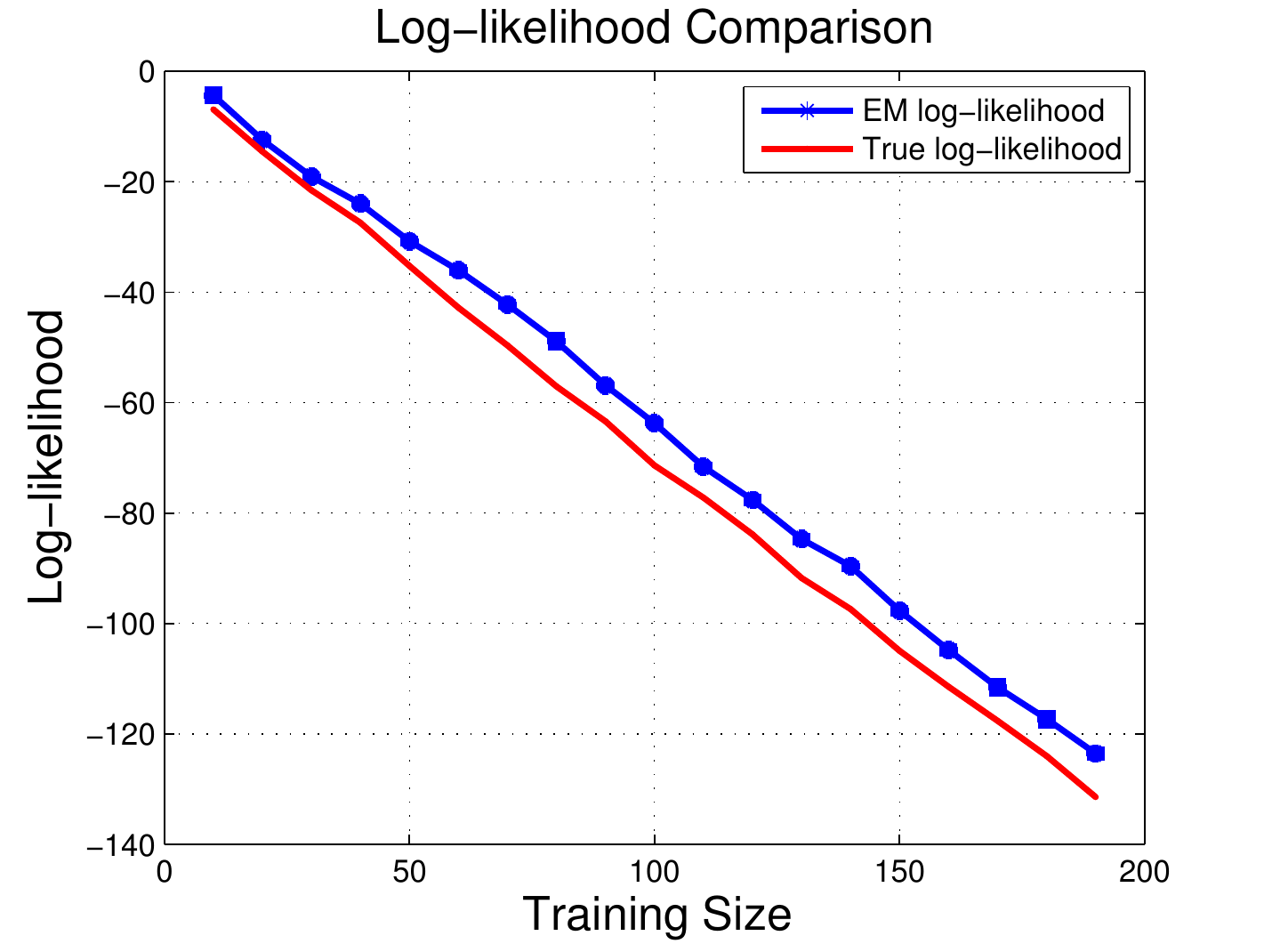}
\caption{Log likelihood comparison.}
\label{fig:em-consistency}
\end{figure}

To test our conjecture that EM is consistent we did another experiment with an HMM of 2 states, 2 observations and 4 parameters. Fig.~\ref{fig:em-consistency} shows the likelihood of the training data for the solutions found by EM in comparison to the true parameters. Our conjecture would not hold if EM found solutions with lower likelihood than for the true parameters because this would mean that it got stuck in a local optimum. However as the the amount of training data increases EM consistently finds solutions with higher likelihood than for the true parameters and the variance of the likelihood vanishes. This suggests that EM found solutions that are all equivalent (i.e. no local optimum that is worse than the other optima).  The fact that the likelihood is higher than for the true parameters simply indicates overfitting. While we suspect that EM is consistent in some settings under suitable conditions (that remain to be proven formally), we note that EM is inconsistent for some \textit{continuous} observation models such as HMMs with continuous observations and mixture of Gaussians (MoGs). For MoGs, it is will known that EM may converge to a mixture of a Dirac distribution centered at one data point with a widespread Gaussian that fits the rest of the data~\cite{bishop2006pattern}. This singular solution has infinite data likelihood, but it does not correspond to the true parameters, confirming the inconsistency of EM.

\section{Conclusion}
Spectral learning is an exciting and promising line of research.  In this work we showed that there is an important gap between the theory and the practice.  We highlighted several open problems regarding negative probabilities and conjectured that EM may be consistent in some settings.

\newpage

\bibliography{reference}

\begin{thebibliography}{10}
\providecommand{\natexlab}[1]{#1}
\providecommand{\url}[1]{\texttt{#1}}
\expandafter\ifx\csname urlstyle\endcsname\relax
  \providecommand{\doi}[1]{doi: #1}\else
  \providecommand{\doi}{doi: \begingroup \urlstyle{rm}\Url}\fi

\bibitem[Anandkumar et~al.(2012{\natexlab{a}})Anandkumar, Foster, and
  Hsu]{Anandkumar}
Anandkumar, Anima, Foster, Dean~P, and Hsu, Daniel.
\newblock {A Spectral Algorithm for Latent Dirichlet Allocation}.
\newblock In \emph{NIPS}, pp.\  926----934, 2012{\natexlab{a}}.

\bibitem[Anandkumar et~al.(2012{\natexlab{b}})Anandkumar, Ge, Hsu, Kakade, and
  Telgarsky]{Anandkumar2012}
Anandkumar, Anima, Ge, Rong, Hsu, Daniel, Kakade, Sham~M, and Telgarsky, Matus.
\newblock {Tensor Decompositions for Learning Latent Variable Models}.
\newblock In \emph{arXiv preprint arXiv:1210.7559}, pp.\  1--55,
  2012{\natexlab{b}}.

\bibitem[Anandkumar et~al.(2012{\natexlab{c}})Anandkumar, Hsu, and
  Kakade]{Anandkumar2012a}
Anandkumar, Animashree, Hsu, Daniel, and Kakade, Sham~M.
\newblock {A Method of Moments for Mixture Models and Hidden Markov Models}.
\newblock \emph{arXiv preprint arXiv:1203.0683}, 2012{\natexlab{c}}.

\bibitem[Bishop(2006)]{bishop2006pattern}
Bishop, Christopher~M.
\newblock \emph{Pattern recognition and machine learning}, volume~1.
\newblock springer New York, 2006.

\bibitem[Boots et~al.(2011)Boots, Siddiqi, and Gordon]{boots2011closing}
Boots, Byron, Siddiqi, Sajid~M, and Gordon, Geoffrey~J.
\newblock Closing the learning-planning loop with predictive state
  representations.
\newblock \emph{The International Journal of Robotics Research}, 30\penalty0
  (7):\penalty0 954--966, 2011.

\bibitem[Casella \& Berger(1990)Casella and Berger]{casella1990statistical}
Casella, George and Berger, Roger~L.
\newblock \emph{Statistical inference}, volume~70.
\newblock Duxbury Press Belmont, CA, 1990.

\bibitem[Hsu et~al.(2012)Hsu, Kakade, and Zhang]{Hsu2012}
Hsu, Daniel, Kakade, Sham~M., and Zhang, Tong.
\newblock {A spectral algorithm for learning Hidden Markov Models}.
\newblock \emph{Journal of Computer and System Sciences}, 78\penalty0
  (5):\penalty0 1460--1480, September 2012.

\bibitem[Parikh \& Xing(2011)Parikh and Xing]{Parikh2011}
Parikh, Ankur and Xing, Eric~P.
\newblock {A Spectral Algorithm for Latent Tree Graphical Models}.
\newblock In \emph{Proceedings of the 28th International Conference on Machine
  Learning}, pp.\  1065--1072, 2011.

\bibitem[Parikh et~al.(2012)Parikh, Teodoru, Tech, Ishteva, and
  Xing]{Parikh2012}
Parikh, Ankur, Teodoru, Gabi, Tech, Georgia, Ishteva, Mariya, and Xing, Eric~P.
\newblock {A Spectral Algorithm for Latent Junction Trees}.
\newblock \emph{arXiv preprint arXiv:1210.4884}, 2012.

\bibitem[Vavasis(2009)]{vavasis2009complexity}
Vavasis, Stephen~A.
\newblock On the complexity of nonnegative matrix factorization.
\newblock \emph{SIAM Journal on Optimization}, 20\penalty0 (3):\penalty0
  1364--1377, 2009.

\end{thebibliography}
\bibliographystyle{icml2014}

\end{document}